\documentclass[nohyperref]{article}

\usepackage{microtype}
\usepackage{graphicx}
\usepackage{subcaption}
\usepackage{booktabs} 

\usepackage{hyperref}

\usepackage[preprint]{icml2022}

\usepackage{amsmath}
\usepackage{amssymb}
\usepackage{mathtools}
\usepackage{amsthm}

\usepackage[capitalize,noabbrev]{cleveref}

\theoremstyle{plain}

\theoremstyle{definition}

\theoremstyle{remark}

\usepackage[textsize=tiny]{todonotes}
\usepackage{bibentry}
\usepackage{multirow}
\usepackage{bm}

\icmltitlerunning{Task-Informed Meta-Learning for Agriculture}

\begin{document}

\twocolumn[
\icmltitle{TIML: Task-Informed Meta-Learning for Agriculture}




\begin{icmlauthorlist}
\icmlauthor{Gabriel Tseng}{nasa,mila}
\icmlauthor{Hannah Kerner}{nasa,umd}
\icmlauthor{David Rolnick}{mila}
\end{icmlauthorlist}

\icmlaffiliation{nasa}{NASA Harvest}
\icmlaffiliation{umd}{University of Maryland, College Park}
\icmlaffiliation{mila}{McGill University and Mila -- Quebec AI Institute}

\icmlcorrespondingauthor{Gabriel Tseng}{gabriel.tseng@mail.mcgill.ca}

\icmlkeywords{Machine Learning, ICML}

\vskip 0.3in
]



\printAffiliationsAndNotice{} 

\begin{abstract}
Labeled datasets for agriculture are extremely spatially imbalanced. When developing algorithms for data-sparse regions, a natural approach is to use transfer learning from data-rich regions. While standard transfer learning approaches typically leverage only direct inputs and outputs, geospatial imagery and agricultural data are rich in metadata that can inform transfer learning algorithms, such as the spatial coordinates of data-points or the class of task being learned. We build on previous work exploring the use of meta-learning for agricultural contexts in data-sparse regions and introduce task-informed meta-learning (TIML), an augmentation to model-agnostic meta-learning which takes advantage of task-specific metadata. We apply TIML to crop type classification and yield estimation, and find that TIML significantly improves performance compared to a range of benchmarks in both contexts, across a diversity of model architectures. While we focus on tasks from agriculture, TIML could offer benefits to any meta-learning setup with task-specific metadata, such as classification of geo-tagged images and species distribution modelling. 
\end{abstract}

\section{Introduction}

Machine learning is useful for inferring comprehensive geospatial information from sparsely labelled data. This is applicable to a wide range of uses, from vegetation height mapping \cite{LANG2019111347} to building footprint detection \cite{building_footprint}. In particular, learning from geospatial data is crucial to better understanding, mitigating, and responding to climate change, with applications ranging from hurricane forecasting \cite{boussioux2021hurricane} to methane detection \cite{methane}. Geospatial data is especially useful to better understand agricultural practices; data such as agricultural land use or yield is extremely incomplete, especially when considered on a global scale, and machine learning is critical in helping fill the gaps in the data. A complete picture of global agricultural practices is vital to mitigate and adapt to the effects of climate change, including by assessing food security in the event of extreme weather, more rapidly responding to food crises, and increasing productive land without sacrificing carbon sinks.

Certain parts of the world collect plentiful field-level agricultural data, but many regions are extremely data-sparse (with this data imbalance reflecting a eurocentric and amerocentric bias as in other labeled datasets in machine learning \cite{geodiversity}). While previous work has investigated transfer learning from data-rich areas to improve performance in data-sparse areas \cite{transferyield,maml_landcover}, geospatial datasets (and agricultural data in particular) are rich in metadata that can inform transfer learning algorithms by enabling models to learn useful context between datapoints, such as the relative geographic locations of datapoints or the higher-level category of the class label \cite{convstar}.

We propose a new method for passing such auxiliary information to the model to improve overall performance and equitable generalization. Specifically, we build on previous work applying Model-Agnostic Meta-Learning (MAML) \cite{maml} to geospatial data. Meta-learning aims to learn a model that can quickly learn a new task from a small amount of new data by optimizing over many training tasks. When using geospatial data, tasks are created by partitioning samples based on agro-ecological \cite{maml_landcover} or political \cite{tseng2021learning,cropharvest} boundaries.

We summarize the main contributions of this paper below\footnote{See \href{https://github.com/nasaharvest/timl}{\texttt{github.com/nasaharvest/timl}} for code \& data}:
\begin{itemize}
    \item We introduce Task-Informed Meta-Learning (TIML), an algorithm designed to augment MAML by incorporating task metadata and removing memorized tasks.
    \item We show that TIML improves performance for both regression (yield estimation) and classification (crop type classification) tasks across a diversity of neural network architectures.
    \item We highlight TIML's ability to learn from very few positive labels and to perform well on tasks where other transfer-learned models do poorly.
\end{itemize}

While we motivate our method and focus our experiments on agricultural applications, we highlight that TIML is not specific to agriculture and could be applied to any meta-learning problem that includes task-specific metadata, such as classification of geo-tagged images \cite{geo_priors_iccv19} or species distribution modelling \cite{species_modelling}.

\section{Related Work}

\begin{figure}
    \centering
    \includegraphics[width=\linewidth]{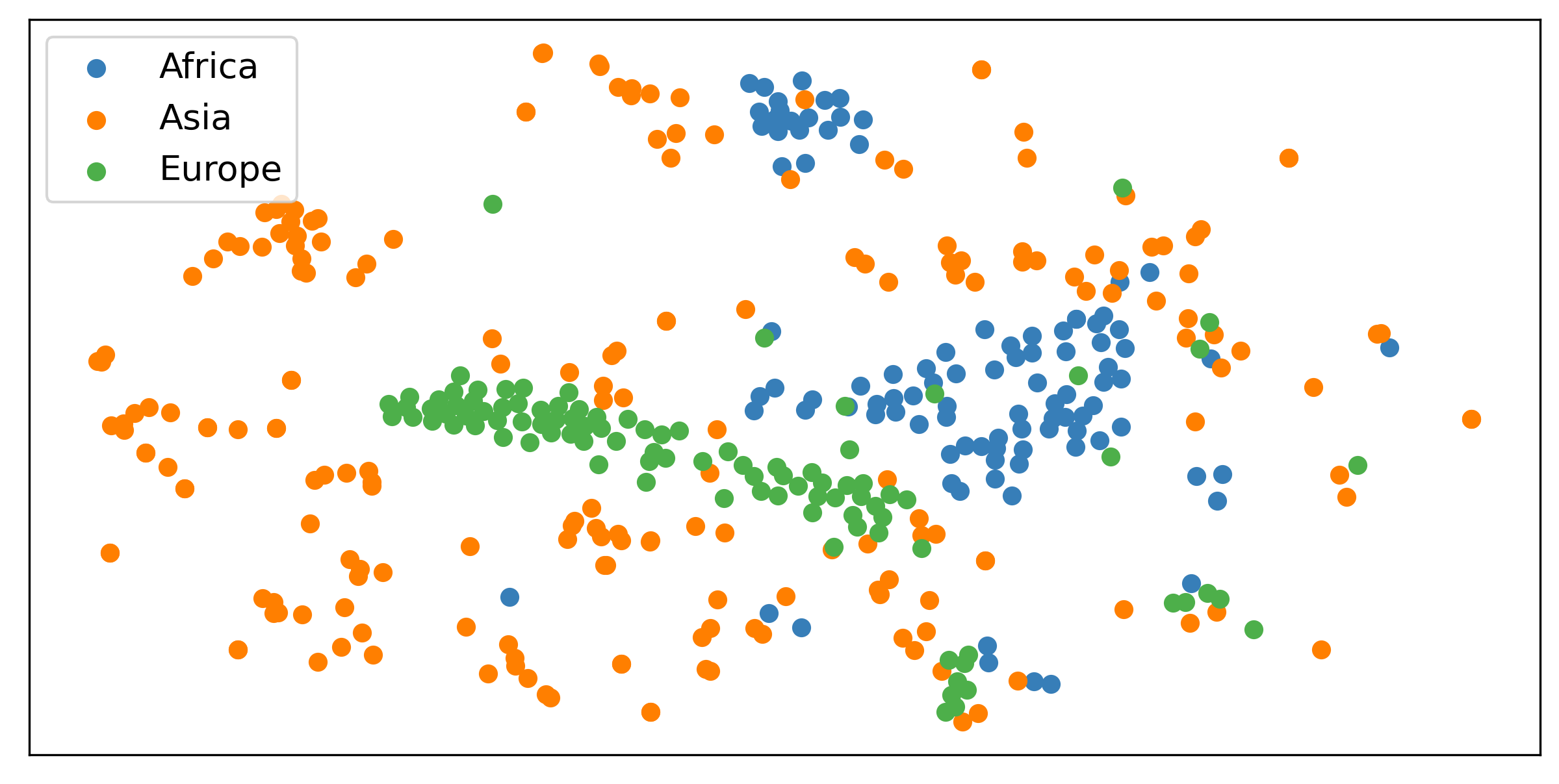}
    \caption{Remote sensing data from the CropHarvest dataset \cite{cropharvest}, collapsed to two dimensions using t-SNE \cite{tsne} and colored according to the continent in which the datapoint is located. Datapoints appear to cluster according to their continent, suggesting that datapoints from the same geographic region share similarities that could be learned by a model. This clustering provides the intuition for the TIML method (that nearby tasks will be more informative during fine-tuning than far-away tasks).}
    \label{fig:clusters}
\end{figure}

\subsection{Transfer learning for remote sensing}
In prior work using machine learning for remote sensing data, there have been numerous efforts to learn from data-rich geographies and transfer the resulting model to data-sparse regions or underrepresented classes. \citet{transferyield} found that training a yield estimation model using data from Argentina boosted the performance of that model in Brazil. In other studies, the source and target tasks are not geographically defined. For example, \citet{poverty_mapping} used transfer learning to improve performance on a data-sparse task (wealth estimation) by first learning a data-rich task (nighttime light estimation).

Prior work has also used multi-task learning to improve model performance for data-sparse tasks. \citet{rapidresponse} trained a multi-task model where one task classifies crops in the data-sparse target region (Togo) and the second task classifies crops elsewhere in the world, and showed that augmenting the data-sparse task with global data improved the model performance in Togo. \citet{chang2019chimera} trained a multi-task neural network to simultaneously classify forest cover type and regress forest structure variables such as biomass to improve model performance on both tasks.

\citet{maml_landcover} introduced the idea of geographically defined tasks for meta-learning and applied it to remote sensing data (specifically for land cover classification). \citet{tseng2021learning} and \citet{cropharvest} used meta-learning for agricultural land cover classification.

These approaches often fail to capture important metadata and expert knowledge about the data-sparse tasks of focus, such as their geographic location relative to the pre-training data or the high-level category of crops being classified. This metadata can be useful for learning relationships between samples or tasks that can improve classification performance---for example, remote sensing observations of maize will be more similar between Kenya and Mali than between Kenya and France (Figure \ref{fig:clusters}). 
In this work, we consider the metadata inherent to agricultural classification tasks, such as the spatial relations between tasks, and how this can inform the model's predictions.

\subsection{Meta-learning}

Meta-learning, or \textit{learning to learn}, consists of learning a function for a task given other example tasks \cite{learning_to_learn}. Recent work in this area has focused on few-shot learning, i.e., learning a function for a task given few training datapoints \cite{prototypical_networks,Sachin2017}. In particular, model-agnostic meta-learning (MAML) \cite{maml} is a few-shot meta-learning algorithm that uses the other example tasks to learn a set of initial weights that can rapidly generalize to a new task. MAML can be used with any neural network architecture.

A more complex variant to few-shot generalization is few-shot \textit{dataset} generalization, where a single model is used to learn from multiple datasets (as opposed to tasks, which can be drawn from the same dataset) \cite{metadataset}. In this regime, one solution is to learn an encoder which can modulate the distribution of weights learned by MAML depending on the dataset being considered \cite{vuorio2019multimodal,pmlr-v139-triantafillou21a}.

In this work, we consider how the MAML weights may be modulated even when all the tasks are drawn from the same dataset. In particular, we consider the special case when there is task-level metadata that can inform the modulation of the MAML weights for specific tasks.

\section{Methods}

\begin{figure}
  \centering
  \includegraphics[width=\linewidth]{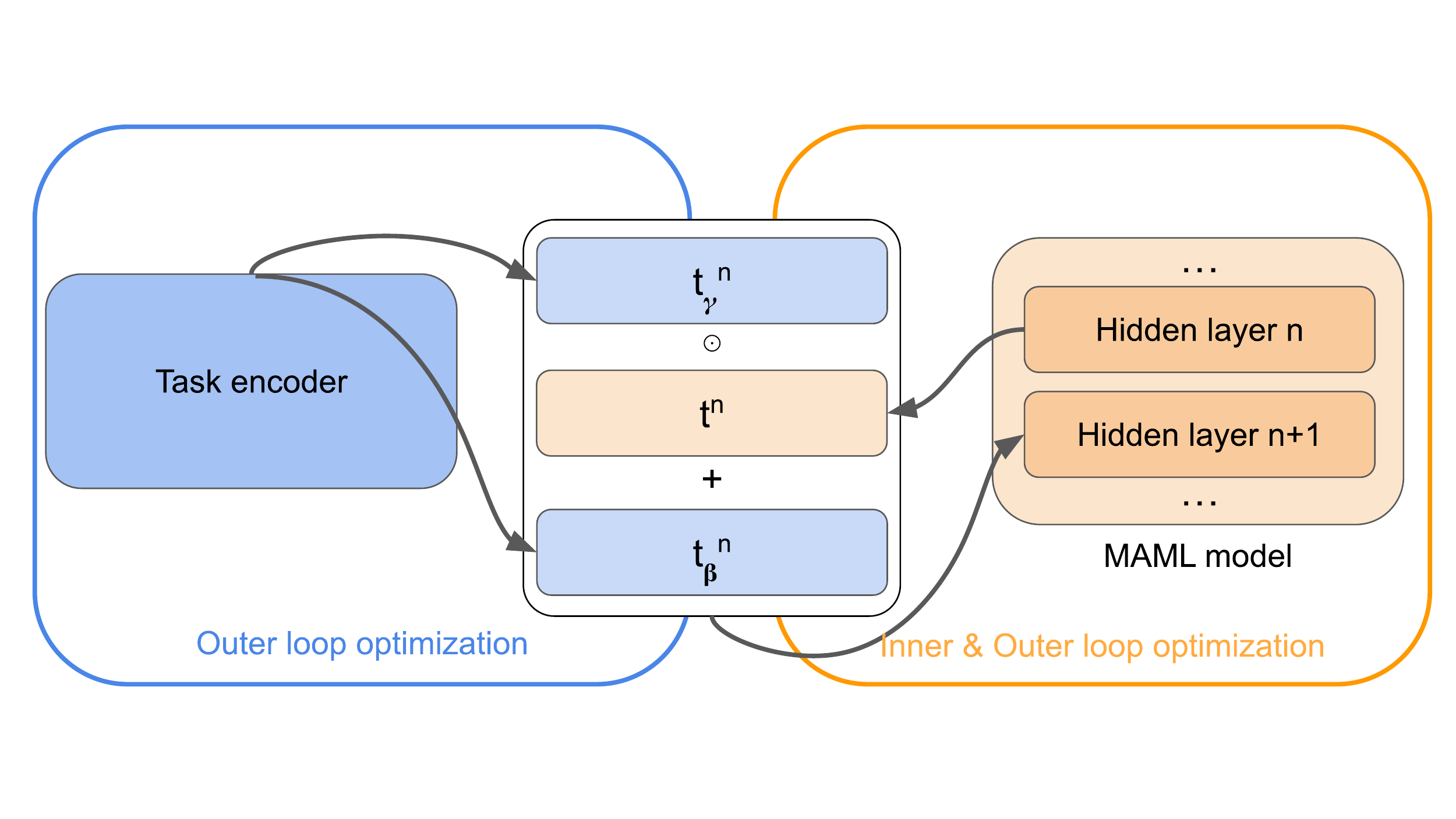}
  \caption{An illustration of the encoder, and the modulation of the MAML learner's hidden vectors using the encoder's output. We highlight the differing optimization regimes for the encoder and the MAML learner -- the encoder's output remains static through the MAML learner's inner loop optimization.}
  \label{fig:maml_diagram}
\end{figure}

Our proposed approach, Task-Informed Meta-Learning (TIML), builds on Model-Agnostic Meta-Learning (MAML) \cite{maml}. Specifically, we modulate the meta-weights learned by MAML depending on the task of interest. This allows different weight initializations to be learned depending on the tasks, allowing the model to learn a wide distribution of tasks.
Model-Agnostic Meta-Learning (MAML) learns a set of model weights $\theta$ which is close to optimal for each of a variety of different tasks, allowing the optimal weights for a specific task to be reached with little data and/or few gradient steps. These initial weights $\theta$ are updated by fine-tuning them on a training task (inner loop training), yielding updated weights $\theta'$. A gradient for $\theta$ is then computed with respect to the loss of the updated model, $L_{\theta '}$. 
This gradient is then used to update $\theta$ (outer loop training).

In the following sections, we first describe how we implemented MAML in a geospatial context (Section \ref{sec:geomaml}) (focusing on how tasks are constructed) before describing the TIML method in more detail (Section \ref{sec:timl}).

\subsection{MAML in a geospatial context} \label{sec:geomaml}
As in previous work applying meta-learning to geospatial data \cite{cropharvest,Wang2020}, we define tasks spatially. Specifically, given a particular task we use political boundaries (counties or countries) to  separate a single dataset into many different tasks. 
The intuition for this is that agricultural practices (and land use) are influenced by the policies and cultural practices of a region, which are often defined along political boundaries (e.g., teff is a popular crop grown in Ethiopia and Eritrea but not bordering countries). This makes political territories useful units when conducting spatial analysis of a region (e.g.~\cite{Kimenyi2014}).
In addition, defining tasks in this way allows for region-specific tasks to be defined. For example, some crops may only be grown in certain parts of the world (e.g., cacao is typically grown within 20 degrees of the equator), or data collection efforts for specific crops may only have occurred in certain areas. Spatially defined tasks mean that the model can be trained to identify these regional crops when it is looking at that region, and not elsewhere.

\subsection{Task-Informed Meta-Learning} \label{sec:timl}

We build on model-agnostic meta-learning \cite{maml}, considering the case where there is additional task-specific information that could inform the model, such as the spatial relationships between tasks. Information such as the spatial coordinates of a task remains static for all datapoints in the task, so is not useful to differentiate positive and negative instances within tasks. However, it may be useful to condition the model prior to inner loop training.

\begin{algorithm}
\caption{Task-Informed Meta-Learning}
\label{alg:maml}
    \begin{algorithmic}[1]
        \STATE \textbf{Require:} $p(\mathcal{T})$: Distribution over tasks
        \STATE \textbf{Require:} $\alpha$, $\beta$: step size
        hyperparameters
        \STATE randomly initialize meta model $\theta_{m}$, task encoder $\theta_{e}$
        \WHILE {not done} 
            \STATE Sample batch of tasks $\mathcal{T}_{i} \sim p(\mathcal{T})$ with task information $t_{i}$
            \FORALL{$\mathcal{T}_{i}, t_{i}$} 
                \STATE Generate task embeddings $\mu_{i} = f(t_{i}; \theta_{e})$
                \STATE Evaluate $\nabla_{\theta_{m}}\mathcal{L}_{\mathcal{T}_{i}}(f_{\theta_{m}}, \mu_{i})$ with respect to K examples
                \STATE Compute adapted meta parameters with gradient descent: $\theta_{m_i}^{'} \leftarrow \theta_{m} - \alpha \nabla_{\theta_{m}}\mathcal{L}_{\mathcal{T}_{i}}(f_{\theta_m}, \mu_{i})$
            \ENDFOR
        \STATE Update $\theta_m \leftarrow \theta_m - \beta \nabla_{\theta_m}\Sigma_{\mathcal{T}_{i} \sim p(\mathcal{T})} \mathcal{L}_{\mathcal{T}_{i}}(f_{\theta_{m_i}^{'}}, \mu_{i})$
        \STATE Update $\theta_{e} \leftarrow \theta_m - \beta \nabla_{\theta_e} \Sigma_{\mathcal{T}_{i} \sim p(\mathcal{T})} \mathcal{L}_{\mathcal{T}_{i}}(f_{\theta_{m_i}^{'}}, \mu_{i})$
        \ENDWHILE
    \end{algorithmic}
    \label{algo:taml}
\end{algorithm}

We introduce Task-Informed Meta-Learning (TIML) (Algorithm \ref{algo:taml}), which modulates the hidden vectors in the meta-model based on embeddings calculated using task information. We encode the task-specific information into a set of vectors -- two for each hidden layer to be modulated in the meta-model, $\bm{t}^i_{\gamma}$ and $\bm{t}^i_{\beta}$. We use feature-wise linear modulation (FiLM \cite{perez2018film}) to modulate the hidden vector outputs of the meta-model using these task encodings. Given a hidden vector output $\bm{h}$, we compute the Hadamard product of $\bm{t}^i_{\gamma}$ and $\bm{h}^i$ and add $\bm{t}^i_{\beta}$ to calculate the modulated vector which is passed to the next layer in the network:
\begin{equation}
    \bm{h}^i_{out} = (\bm{t}^i_{\gamma} \odot h^i) + \bm{t}^i_{\beta}
\end{equation}

These task embeddings are updated in the outer loop during training. This means that when the meta-model is being fine-tuned for a specific task, the embeddings remain constant for all datapoints in that task. We illustrate this in Figure \ref{fig:maml_diagram}.

\paragraph{Task encoder} We use a task encoder to learn the embeddings. This encoder consists of linear blocks, where each block contains a linear layer with a GeLU activation \cite{gelu}, group normalization \cite{groupnorm} and dropout \cite{srivastava2014dropout}. The task information is encoded into a hidden task vector. Independent blocks are then used to generate an embedding for each hidden vector in the classifier to be modulated. 

\subsection{Forgetful Meta-Learning}

Due to the spatial imbalance of data globally \cite{Vries_2019_CVPR_Workshops,geodiversity}, spatially partitioned meta-learning tasks may not be geographically well distributed. They can also be semantically imbalanced. In the CropHarvest dataset \cite{cropharvest}, a large fraction of the tasks are crop vs.~non-crop tasks, reflecting the large number of binary crop vs.~non-crop datapoints in the dataset (65.8\% of all instances only have crop vs.~non-crop labels). We find that in such settings the model is liable to memorize many similar tasks to the detriment of its ability to learn more difficult or rarer tasks, thus hurting generalization performance for the fine-tuning tasks. This limitation is not limited to geospatial or agricultural datasets and could occur for any dataset with imbalanced classes or task difficulty. 

Although complex meta-learning methods have been designed to optimize for performance on highly challenging tasks \cite{Jamal_2019_CVPR,NEURIPS2020_da8ce53c}, we take advantage of the large number of similar tasks in the geospatial data setting to introduce a simple method to prevent memorization of certain tasks: removing training tasks the model has memorized, where memorization is defined as having exceeded a performance threshold for a task over a continuous set of epochs. We call this method ``forgetfulness.'' Reducing the training set size before training has been previously explored \cite{Ohno-Machado1998,8890886} to reduce training time - we do it dynamically to improve performance.

\section{Datasets}

TIML is designed for transfer learning regimes which can be structured in terms of meta-learning (learning from many different tasks). In addition, TIML expects metadata that inform how the model should consider each task and remain constant for all datapoints within the task. 
We consider two datasets well suited to this regime.
We consider a regression and a classification task to demonstrate the suitability of TIML in both contexts.

\subsection{Crop Type Classification} \label{sec:crop_classification}

Up to date cropland maps are critical to understanding the climate impacts of agriculture \cite{Song2021}. Crop type classification consists of predicting whether or not a given instance contains a crop of interest. Specifically, given a remote sensing-derived pixel time series for a specific latitude and longitude and a crop of interest, the goal is to output a binary value describing whether the crop of interest is being grown at that pixel location.

\subsubsection{Data description}
We use the CropHarvest dataset \cite{cropharvest}. This dataset consists of 90,480 globally distributed datapoints with the associated satellite pixel time series for each point. Of these datapoints, 30,899 (34.2\%) contain multi-class agricultural labels; the remaining datapoints contain  binary ``crop'' or ``non-crop'' labels. Each datapoint is accompanied by a pixel time series from 4 remote sensing products: Sentinel-2 L1C optical observations, Sentinel 1 synthetic aperture radar observations, ERA5 climatology data (precipitation and temperature), and topography (slope and elevation) from a Digital Elevation Model (DEM). The time series includes 1 year of data at monthly timesteps.

\subsubsection{Task construction for meta-learning}
As with the CropHarvest benchmarks, we defined tasks spatially using bounding boxes for countries drawn by Natural Earth \cite{naturalearth}. Tasks consist of binary classification of pixels as either crop vs.~non-crop or a specific crop type vs.~rest. This yielded 525 tasks, which were randomly split into training and validation tasks. Three evaluation tasks (described in Section \ref{sec:cropharvest_eval}) were withheld from the initial training. For each evaluation task, we fine-tuned the model on that task's training data before evaluating the model on that task's test data.

\paragraph{Task Information}
Task information is encoded in a 13-dimensional vector. Three dimensions are used to encode spatial information, consisting of latitude and longitude transformed to $[cos(\textrm{lat}) \times cos(\textrm{lon}), cos(\textrm{lat}) \times sin(\textrm{lon}), sin(\textrm{lat})]$.
This transforms the spatial information from spherical to Cartesian coordinates, ensuring that transformed values at the extreme longitudes are close to each other. The remaining 10 dimensions are used to communicate the type of task the model is being asked to learn. This consists of a one-hot encoding of crop categories from the UN Food and Agriculture Organization (FAO) indicative crop classification \cite{fao_classification}, with an additional class for non-crop. For crop vs.~non-crop tasks, positive examples are given the value $\frac{1}{n}$ across all the $n=9$ crop type categories.

\subsubsection{Evaluation}\label{sec:cropharvest_eval}

The CropHarvest dataset is accompanied by 3 evaluation tasks which test the ability of a pre-trained model to learn from a small number of in-distribution datapoints in a variety of agroecologies. These test tasks cover a variety of agroecologies. We describe each task and the accompanying training data below. 

\textbf{Togo crop vs.~non-crop}: The goal of this task is to classify datapoints as crop or non-crop in Togo. The training set consists of 1,319 datapoints and the test set consists of 306 datapoints -- 106 (35\%) positive and 200 (65\%) negative --  sampled from random locations within the country. 

The two other evaluation tasks consist of classifying a specific crop. Thus, ``rest'' below includes all other crop and non-crop classes. For both tasks, entire polygons delineating a field (as opposed to single pixels within a field) were collected, allowing evaluation across the polygons. However, during training, only the polygon centroids were used.

\textbf{Kenya maize vs.~rest}: The training set consists of 1,345 imbalanced samples (266 positive and 1,079 negative samples). The test set consists of 45 polygons containing 575 (64\%) positive and 323 (36\%) negative pixels.

\textbf{Brazil coffee vs.~rest}: The training set consists of 794 imbalanced samples (21 positive and 773 negative samples). The test set consists of 66 polygons containing 174,026 (25\%) positive and 508,533 (75\%) negative pixels.

\subsection{Yield Estimation}

Accurate and timely yield estimates are a key input to food security forecasts \cite{BeckerReshef2019}, and to better understand how food production can be sustainably managed \cite{Lark2020}. Yield estimation is a regression task which consists of estimating the yield -- the amount of crop harvested per unit of land -- of an area, given remote sensing data of that area. Specifically, we estimate soybean yield in the top soybean-producing states in the United States.

\subsubsection{Data description} \label{sec:yield_data}
We recreate the yield prediction dataset originally collected by \citet{You_Li_Low_Lobell_Ermon_2017}. This dataset consists of county-level soybean yields for the 11 US states accounting for over 75\% of national soybean production from 2009 to 2015. MODIS reflectance \cite{MODISTERRA} and temperature data \cite{MODISTEMP} are used to construct the remote sensing inputs. Since counties cover large areas, inputting the raw satellite data to the model would create extremely high-dimensional inputs. To handle this, \citet{You_Li_Low_Lobell_Ermon_2017} assumed \textit{permutation invariance}, meaning the positions of farmland pixels in a county do not affect yield, since they only indicate the positions of cropland. This allows all cropland pixels in a county (based on the MODIS land cover map \cite{FRIEDL2010168}) to be mapped to a histogram of pixel values, significantly reducing the dimensionality of the input. This is the predictand for soybean yields in each county.
Since the original \citet{You_Li_Low_Lobell_Ermon_2017} paper was released, the MODIS data product version has incremented from version $5.1$ to $6.0$. Therefore, our histograms are similar but not identical to those in \citet{You_Li_Low_Lobell_Ermon_2017}. We note that this dataset was also released by \cite{yeh2021sustainbench} but we chose to recreate the dataset from \citet{You_Li_Low_Lobell_Ermon_2017} since we wanted to maintain the temporal validation originally used (as opposed to the random split used in \citet{yeh2021sustainbench}).

\subsubsection{Task construction for meta-learning}
We define tasks to be individual counties, with task $(X,y)$ pairs consisting of histograms and yields for different years.

\paragraph{Task Information}
As with the crop type classification task (Section \ref{sec:crop_classification}), we use 3 dimensions in the task information vector to encode transformed latitude and longitude values describing the location of the county. In addition, we include a one-hot encoding communicating which state the county is in to the model based on the intuition that agricultural practices vary enough across states that it may help the model to have this difference explicitly communicated.

\subsubsection{Evaluation}
We use temporal validation; specifically, for each year in $\{2011, 2012, 2013, 2014, 2015\}$, we train a model using all the data prior to that year, and evaluate the performance of the model for the unseen year.

\section{Experiments}

\subsection{Crop Type Classification}

\begin{table*}
  \caption{Results for the \textbf{crop type classification} evaluation tasks. All results are averaged from 10 runs and reported with the accompanying standard error. We report the area under the receiver operating characteristic curve (AUC ROC) and the F1 score using a threshold of 0.5 to classify a prediction as the positive or negative class. We highlight the {\color{blue} \textbf{first}} and \textbf{second} best metrics for each task. TIML achieves the highest F1 score of any model on the Brazil task and the best AUC ROC and F1 scores when averaged across the 3 tasks. We highlight the improvement of TIML relative to other transfer-learning models, showing it is able to leverage task structure to significantly increase performance on the CropHarvest dataset.}
  \label{tab:benchmark-results}
  \centering
  \begin{tabular}{llrrrr}
    \toprule[1.5pt]
     & Model & Kenya & Brazil & Togo & Mean \\
     \toprule[1.5pt]
     \multirow{8}*{\rotatebox[origin=c]{90}{AUC ROC}} & Random Forest & $0.578 \pm 0.006$ & $0.941 \pm 0.004$ & $0.892 \pm 0.001$ & 0.803 \\
     & No pre-training & $0.329 \pm 0.011$ & $0.898 \pm 0.010$ & $0.861 \pm 0.002$ & 0.700 \\
     & Crop pre-training & $0.694 \pm 0.001$ & $0.820 \pm 0.002$ & \bm{$0.894 \pm 0.000$} & 0.801 \\
     & MAML & \bm{$0.729 \pm 0.001$} & $0.831 \pm 0.005$ & $0.878 \pm 0.001$ & 0.843 \\
     \cmidrule[0.5pt]{2-6}
      & TIML & {\color{blue} \bm{$0.794 \pm 0.003$}} & {\color{blue} \bm{$0.988 \pm 0.001$}} & $0.890 \pm 0.000$ & {\color{blue} \bm{$0.890$}} \\
     & \hspace{5mm} no forgetfulness & $0.779 \pm 0.003$ & $0.877 \pm 0.003$ & $0.893 \pm 0.001$ & 0.850 \\
     & \hspace{5mm} no encoder & $0.712 \pm 0.001$ & \bm{$0.977 \pm 0.002$} & {\color{blue} \bm{$0.895 \pm 0.000$}} & \bm{$0.862$} \\
     & \hspace{5mm} no task info or encoder & $0.690 \pm 0.001$ & $0.977 \pm 0.002$ & $0.876 \pm 0.001$ & 0.848 \\
     \midrule[1.5pt]
    \multirow{8}*{\rotatebox[origin=c]{90}{F1 score}} & Random Forest & $0.559 \pm 0.003$ & $0.000 \pm 0.000$ & \bm{$0.756 \pm 0.002$} & 0.441 \\
    & No pre-training & $0.782 \pm 0.000$ & \bm{$0.764 \pm 0.012$} & $0.720 \pm 0.005$ & \bm{$0.734$} \\
    & Crop pre-training & $0.819 \pm 0.001$ & $0.619 \pm 0.005$ & $0.713 \pm 0.002$ & 0.613 \\
     & MAML & $0.828 \pm 0.001$ & $0.496 \pm 0.001$ & $0.662 \pm 0.001$ & 0.652 \\
    \cmidrule[0.5pt]{2-6}
    & TIML & \bm{$0.838 \pm 0.000$} & {\color{blue} \bm{$0.835 \pm 0.012$}} & $0.732 \pm 0.002$ & {\color{blue} \bm{$0.802$}} \\
    & \hspace{5mm} no forgetfulness & {\color{blue} \bm{$0.840 \pm 0.000$}} & $0.537 \pm 0.002$ & {\color{blue} \bm{$0.764 \pm 0.002$}} & 0.724 \\
    & \hspace{5mm} no encoder & {\color{blue} \bm{$0.840 \pm 0.000$}} & $0.473 \pm 0.002$ & $0.691 \pm 0.001$ & 0.691 \\
    & \hspace{5mm} no task info or encoder & $0.837 \pm 0.001$ & $0.473 \pm 0.001$ & $0.645 \pm 0.002$ & 0.652 \\
    \bottomrule[1.5pt]
  \end{tabular}
\end{table*}

We evaluate TIML by training it on the CropHarvest dataset and fine-tuning it on the evaluation tasks, as was done for the benchmark results released with the dataset in \citet{cropharvest}. MAML (and by extension, TIML) can be applied to any neural network architecture. We use the same base classifier and hyperparameters as in \citet{cropharvest}: a 1-layer LSTM model followed by a linear classifier.

\subsubsection{Ablations}
We perform 3 ablations to understand the effects of different components of TIML on overall model performance:
\begin{itemize}
    \item \textbf{No forgetfulness}: A TIML model trained without forgetfulness; no tasks are removed in the training loop.
    \item \textbf{No encoder}: A TIML model with no encoder. The task information is instead appended to every raw input timestep and passed directly to the classifier.
    \item \textbf{No task information or encoder}: No task information passed to the model at all. This model is effectively a normal MAML model, trained with forgetfulness.
\end{itemize}

\subsubsection{Baselines}\label{sec:baselines}
We compare the TIML architecture to 4 baselines. As with TIML, we fine-tune these models on each benchmark task's training data and then evaluate them on the task's test data:
\begin{itemize}
    \item \textbf{MAML}: A model-agnostic meta-learning classifier without the task information.
    \item \textbf{Crop pre-training}: A classifier pre-trained to classify all data as crop or non-crop (without task metadata), then re-trained on each test task.
    \item \textbf{No pre-training}: A randomly initialized classifier, which is not pre-trained on the global CropHarvest dataset but instead is trained directly on the test task training data.
\end{itemize}
In addition, we trained a \textbf{Random Forest} baseline implemented using scikit-learn \cite{sklearn} with the default hyperparameters.

\subsection{Yield Estimation}

We apply TIML to the original network architectures used by \citet{You_Li_Low_Lobell_Ermon_2017} -- a 1-layer LSTM and a CNN-based regressor. In addition to the remote sensing input, the Deep Gaussian Process baseline model (described in Section \ref{sec:yield_baseline}) receives as input the year of each training point. We therefore append the year to each timestep of the input to the TIML LSTM, so that the model has comparable inputs to the Deep Gaussian Process.
The CNN-based models receive only the remotely sensed data as input.

\subsubsection{Baselines}\label{sec:yield_baseline}

We compared TIML to 2 baselines: the Deep Gaussian Process models (proposed by \citet{You_Li_Low_Lobell_Ermon_2017} alongside the yield estimation dataset) and standard MAML.

\paragraph{Deep Gaussian Process} To train a Deep Gaussian Process, a deep learning model is first trained to estimate yield given the remote sensing dataset described above. The final hidden vector $h(x)$ of the model (for each input) is used as input to a Gaussian process:
\begin{equation}
    y(x) = f(x) + h(x)^T \textrm{ where } f(x) \sim \mathcal{GP}(0, k(x, x'))
\end{equation}
where the kernel function $k$ is conditioned on both the location of the datapoint (defined by its latitude and longitude), $g_{l}$, and the year of the datapoint, $g_{y}$:
\begin{equation}
    \small
    k(x, x') = \sigma^2 \textrm{exp}[-\frac{\lVert g_{l}-g'_{l} \rVert^2_2}{2r_{l}}-\frac{\lVert g_{y}-g'_{y} \rVert^2_2}{2r_{y}}] + \sigma^2_e \delta_{g, g'}
\end{equation}
We include baselines with and without a Gaussian process (i.e.,~using the outputs of the deep learning models directly instead of passing the final hidden vectors to a Gaussian process). We note that this implementation of Deep Gaussian Processes differs from \citet{pmlr-v31-damianou13a}.

As noted in Section \ref{sec:yield_data}, the MODIS datasets have been updated since the original Deep Gaussian Process models were run. We therefore retrain them to obtain our baseline results. We use the same hyperparameters as \citet{You_Li_Low_Lobell_Ermon_2017}, with the addition of early stopping when training. \citet{You_Li_Low_Lobell_Ermon_2017}'s original results are included for comparison.

Additional implementation details for the crop classification and yield estimation datasets are available in Appendix \ref{ap:implementation_details}.

\begin{table*}[]
    \centering
    \caption{The RMSE of county-level model performance for the \textbf{yield estimation} task. We use temporal validation to evaluate the model. Specifically, for each year, models are trained with data up to that year and evaluated with that year's data. All models are calculated from an average of 10 runs, with the standard error reported. We highlight the {\color{blue} \textbf{first}} and \textbf{second} best metrics for each task. For completeness, we include the results reported by \cite{You_Li_Low_Lobell_Ermon_2017}, but highlight that these results were obtained on the MODIS 5.1 dataset (whilst all other models were trained on the MODIS 6.0 dataset) and are the result of 2 runs, compared to 10 runs for all other models. TIML improves on the Deep Gaussian Process models for both architectures, even though MAML performs significantly worse than all other models. This suggests that in some cases, the task information is necessary for meta-learning to work.}
    \label{tab:yield_results}
    \begin{tabular}{lrrrrrr}
    \toprule[1.5pt]
    Model & 2011 & 2012 & 2013 & 2014 & 2015 & Mean\\
    \midrule
    LSTM & $5.62 \pm 0.10$ & $6.60 \pm 0.29$ & $5.57 \pm 0.21$ & $6.63 \pm 0.13$ & $6.69 \pm 0.31$ & 6.22 \\
    \hspace{5mm} + GP & $5.32 \pm 0.10$ & \bm{$5.83 \pm 0.18$} & $5.70 \pm 0.19$ & $5.61 \pm 0.12$ & $\bm{5.24 \pm 0.14}$ & $ \bm{5.54}$ \\
    \hspace{5mm} + MAML & $26.90 \pm 0.01$ & $30.97 \pm 0.01$ & $29.57 \pm 0.01$ & $30.84 \pm 0.01$ & $32.02 \pm 0.01$ & 30.06 \\
    \hspace{5mm} + TIML & {\color{blue} \bm{$5.16 \pm 0.03$}} & {\color{blue} \bm{$5.77 \pm 0.05$}} & \bm{$5.39 \pm 0.02$} & $5.24 \pm 0.04$ & {\color{blue} \bm{$4.89 \pm 0.04$}} & {\color{blue} \bm{$5.29$}} \\
    \midrule[1.5pt]
    CNN & $6.08 \pm 0.77$ & $6.94 \pm 1.83$ & $6.42 \pm 1.23$ & {\color{blue} \bm{$4.80 \pm 0.83$}} & $5.57 \pm 0.38$  & 5.96 \\
    \hspace{5mm} + GP & $5.55 \pm 0.14$ & $6.18 \pm 0.49$ & $6.44 \pm 0.67$ & \bm{$4.87 \pm 0.31$} & $6.02 \pm 0.26$  & 5.81 \\
    \hspace{5mm} + MAML & $12.93 \pm 0.05$ & $8.28 \pm 0.07$ & $7.98 \pm 0.04$ & $12.05 \pm 0.05$ & $7.69 \pm 0.06$ & 9.79 \\
    \hspace{5mm} + TIML & \bm{$5.23 \pm 0.02$} & $6.59 \pm 0.02$ & {\color{blue} \bm{$5.34 \pm 0.01$}} & $4.93 \pm 0.02$ & $6.35 \pm 0.01$ & 5.69 \\
    \midrule[1.5pt]
    \cite{You_Li_Low_Lobell_Ermon_2017} & & & & & & \\
    LSTM + GP & 5.77 & 6.23 & 5.96 & 5.70 & 5.49 & 5.83 \\
    CNN + GP & 5.70 & 5.68 & 5.83 & 4.89 & 5.67 & 5.55 \\
    \bottomrule
    \end{tabular}
\end{table*}

\section{Results \& Discussion}

\begin{figure}
    \centering
    \begin{subfigure}[b]{0.75\linewidth}
        \centering
        \includegraphics[width=\textwidth]{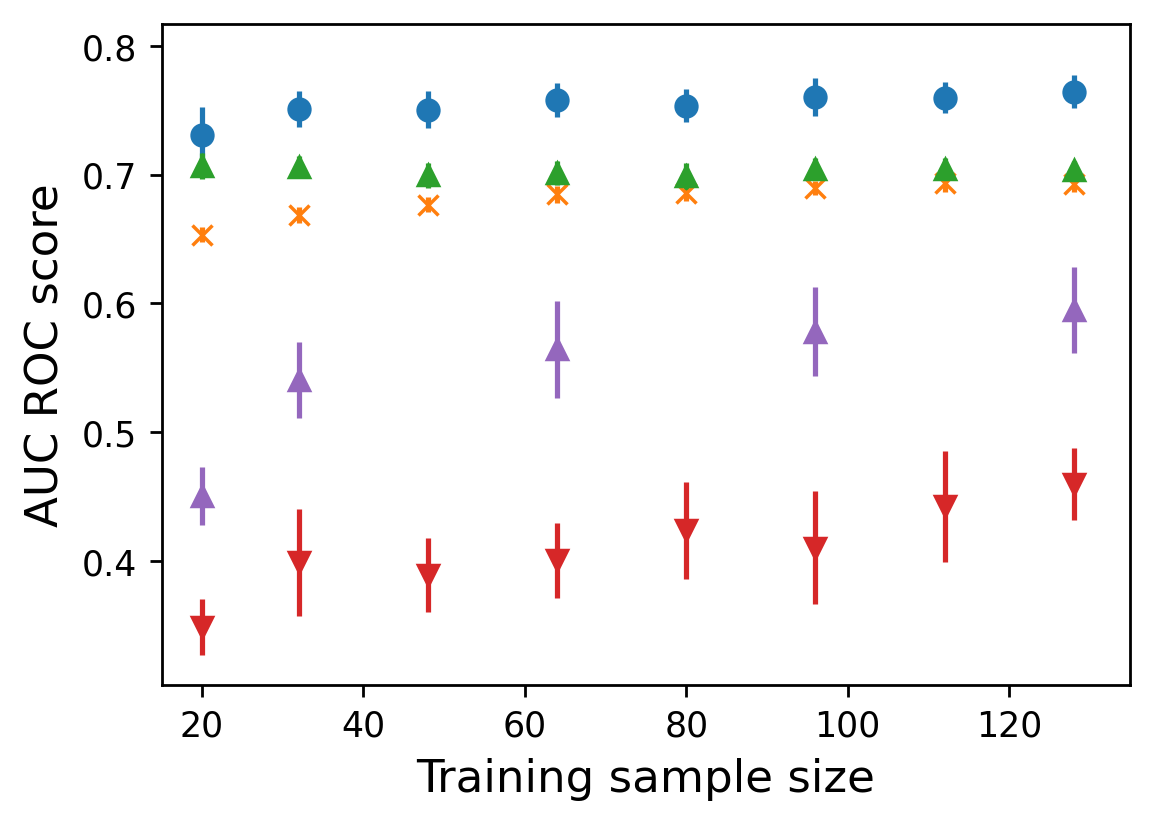}
        \caption{Kenya: Maize vs.~Rest}
        \label{fig:subset_results_kenya}
    \end{subfigure}
        \begin{subfigure}[b]{0.75\linewidth}
        \centering
        \includegraphics[width=\textwidth]{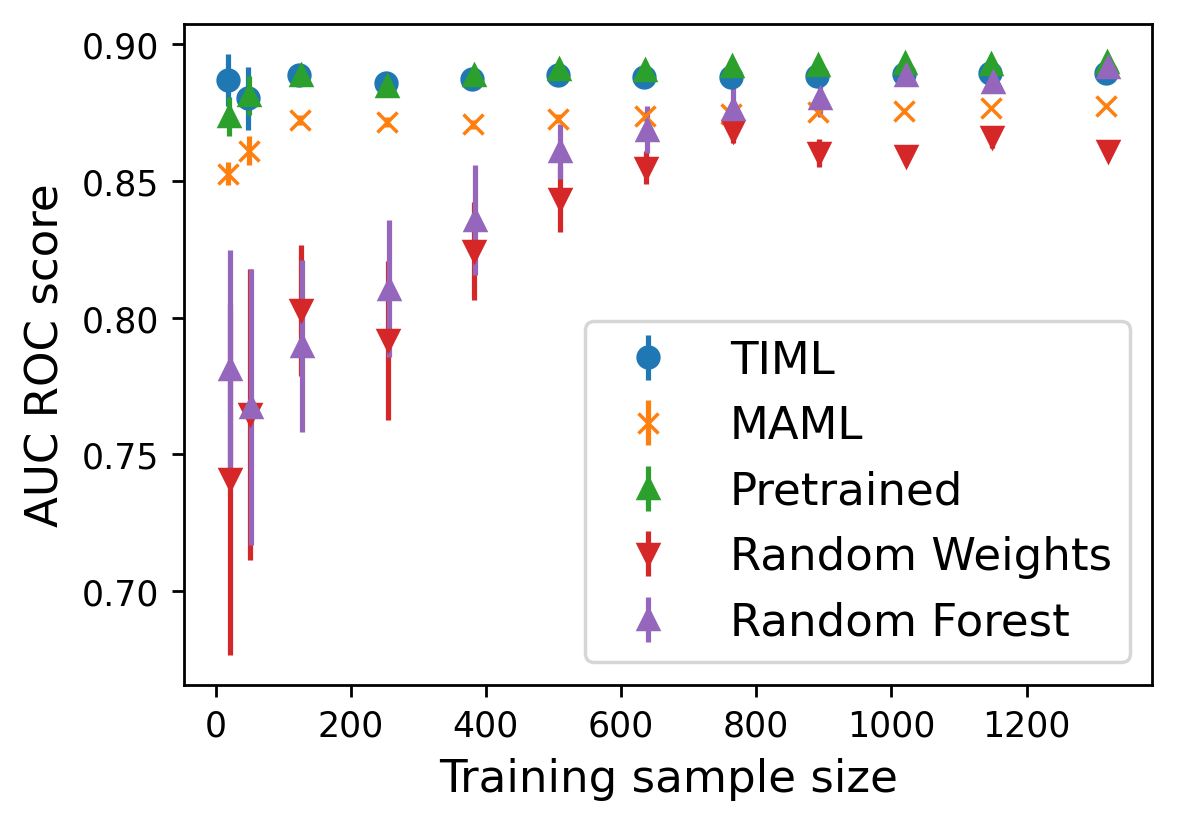}
        \caption{Togo: Crop vs.~Non Crop}
        \label{fig:subset_results_togo}
    \end{subfigure}
    \caption{Results of TIML and the benchmark models when trained on a subset of the evaluation training datasets for the Crop Type Classification Task. Specifically, we plot results for (Figure \ref{fig:subset_results_kenya}) the Kenya Maize vs.~rest evaluation task and (Figure \ref{fig:subset_results_togo}) Togo Crop vs.~Non Crop evaluation task. All results are averaged from 10 runs, and reported with standard error bars. For both tasks, the subset is balanced so that it contains an equal number of positive and negative samples. TIML is the best performing model in Kenya, and - alongside the crop vs. non crop pretrained model - is among the best performining models in Togo for all subset sizes, indicating TIML's ability to learn from limited datast sizes. We highlight that for the smallest training sample size, consisting of 20 samples, TIML is the best performing algorithm.}
    \label{fig:subset_results}
\end{figure}

\subsection{Crop Type Classification}
We show the model results for TIML, its ablations and all baseline models when trained on the CropHarvest dataset in Table \ref{tab:benchmark-results}. Like \citet{cropharvest}, we report the AUC ROC score and the F1 score calculated using a threshold of 0.5. Overall, TIML is the best performing algorithm on the CropHarvest dataset, achieving the highest F1 and AUC ROC scores when averaged across all tasks. TIML is consistently the best performing algorithm on every task. In particular, TIML is the only transfer learning model that outperforms a randomly-initialized model in the challenging Brazil task, where there are only 26 positive datapoints.

\subsubsection{Effects of transfer learning} Standard transfer learning from the global dataset is not guaranteed to confer advantages to the model. For example, first training using MAML or crop pre-training results in lower performance on the Brazil task compared to an LSTM initialized with random weights. We hypothesize this may be due to the difference in distribution of the Brazil task data relative to the other tasks the models are trained on. TIML is the only model to see significant improvements in performance compared to the randomly initialized model, suggesting conditioning the model with prior, domain-specific information about the tasks can help to model the diversity of samples in the CropHarvest dataset.

\subsubsection{Forgetfulness} 
Forgetfulness -- when coupled with task information -- improves model performance in more challenging tasks without penalizing performance elsewhere. Training TIML with forgetfulness significantly boosts performance in the Brazil task without substantially impacting performance on the other tasks, and yields significantly higher mean F1 and AUC ROC scores when measured across all tasks. However, training TIML forgetfully without the task information (TIML with no task information or encoder) yields comparable results to the baseline MAML model trained without forgetfulness. We therefore hypothesize that task information provides useful context around which tasks are being kept and forgotten during training, allowing TIML to learn from more difficult tasks in the ``forgetful'' regime without forgetting easier tasks it has already learned.

\subsubsection{Effect of task information} Including task information in the model improves performance, both when it is concatenated to the input data and when it is passed to the model through TIML. However, there are significant differences in performance depending on \textit{how} this information is passed to the model: passing the task information directly to the classifier (TIML with no encoder) yields mixed results (lower AUC ROC in Kenya, and lower F1 scores in Brazil and Togo compared to MAML or crop-pretraining models). Training TIML with the encoder significantly boosts performance in these regimes, yielding the highest mean AUC ROC and F1 scores. We hypothesize that because the task information remains static for all datapoints in a task, it is challenging for the model to learn from them during the inner loop optimization -- the encoder architecture allows it to only be optimized in the outer loop.

\subsubsection{Effects of fine-tuning dataset sizes}
TIML excels at learning from small dataset sizes. We plot the performance of the models as a function of training set size in Figure \ref{fig:subset_results} for the Kenya and Togo evaluation tasks (the Brazil task has only 26 positive examples, and is therefore already in the small-dataset size regime). In both the Togo and Kenya tasks, the TIML model is amongst the most performant algorithms (as measured by AUC ROC score) for all subset sizes. We highlight that for the smallest sample size (20 fine-tuning samples), TIML is the best performing algorithm for both the Togo and Kenya evaluation tasks.

\subsection{Yield Estimation}
We share yield estimation results in Table \ref{tab:yield_results}. Like \citet{You_Li_Low_Lobell_Ermon_2017}, we report the RMSE score averaged across all counties and use temporal validation to evaluate the models. 
The TIML and MAML LSTM receive the year as input (to reflect the data available to the Deep Gaussian Process), but the TIML and MAML CNN do not.

For both the LSTM and CNN architectures, TIML is the most performant model. This is the case even though the Deep Gaussian Process is much more memory intensive, since it requires all predictions and hidden vectors (for the training and test data) to be computed together for the Gaussian process modelling step; this may be infeasible for larger datasets. TIML requires substantially less memory since it considers each county independently.

It is also worth noting that while TIML achieves the best result of all models, MAML performs significantly worse than all other models. This suggests that in some contexts, the task information is necessary for meta-learning to work.


\section{Conclusion}
In conclusion, we introduce task-informed meta-learning (TIML), a method for conditioning the model with prior information about a specific task. Specifically, the task information is encoded into a set of vectors which are used to modulate the weights learned by a MAML learner prior to task-specific fine-tuning. In addition, we introduce the concept of ``forgetful'' meta-learning, which can improve meta-learning performance when there are many similar tasks to learn from. We apply TIML to a range of tasks (classification and regression) and a range of model architectures (RNNs and CNNs), demonstrating its utility in a variety of regimes (including those with very few data points, and regimes in which standard MAML fails completely). Although we focus on agriculture-related tasks, TIML is not specific to agriculture and can be applied to any meta-learning setup with task-level metadata.



\bibliography{example_paper}
\bibliographystyle{icml2022}

\newpage
\appendix
\section{Implementation Details} \label{ap:implementation_details}
We implement TIML in PyTorch \cite{pytorch}, using the learn2learn library \cite{learn2learn}. All MAML and TIML models are trained using the same optimizer hyperparameters. Specifically, we use an inner loop learning rate of $10^{-4}$. We use an Adam optimizer on the outer loop (for both the classifier and the encoder), with a Cosine Annealing Learning rate (as per \citet{mamlplus}). For both the classifier and encoder, we use an initial learning rate of $10^{-4}$ and a minimum learning rate of $10^{-5}$.

When fine-tuning, we use the same learning rate as the inner loop learning rate ($10^{-4}$) for all models with the exception of the yield-estimation standard-MAML CNN. The standard-MAML CNN experienced an exploding loss using this learning rate, so we reduced the learning rate to $10^{-5}$ when fine-tuning it.

Both MAML and TIML are trained for 1000 epochs - we selected the model checkpoint with the best performance on the validation set (consisting of 10\% of the training tasks, up to a maximum of 50 tasks).

\begin{figure}
    \centering
    \begin{subfigure}[b]{0.3\linewidth}
        \centering
        \includegraphics[width=\textwidth]{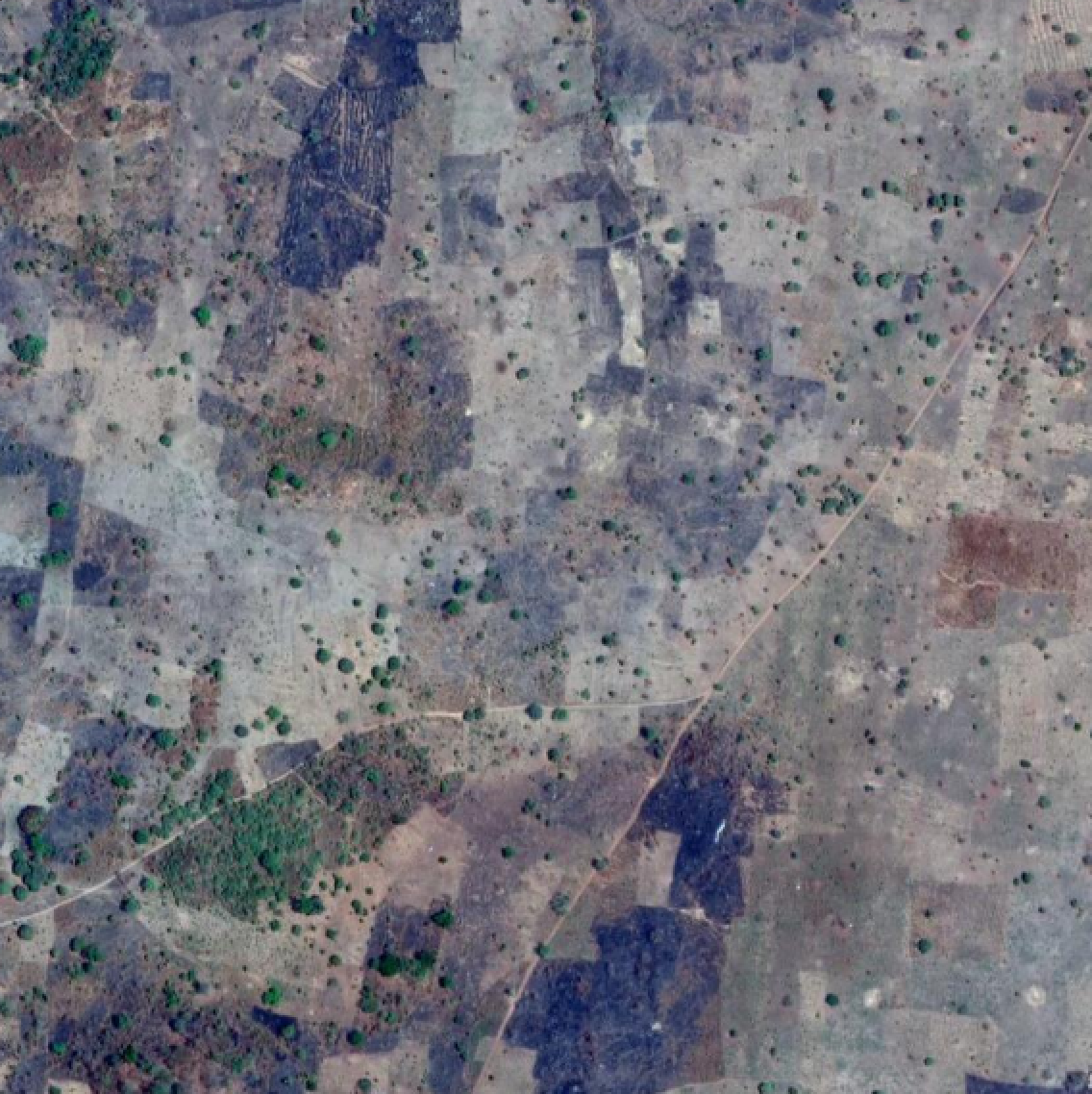}
        \caption{Togo}
    \end{subfigure}
        \begin{subfigure}[b]{0.3\linewidth}
        \centering
        \includegraphics[width=\textwidth]{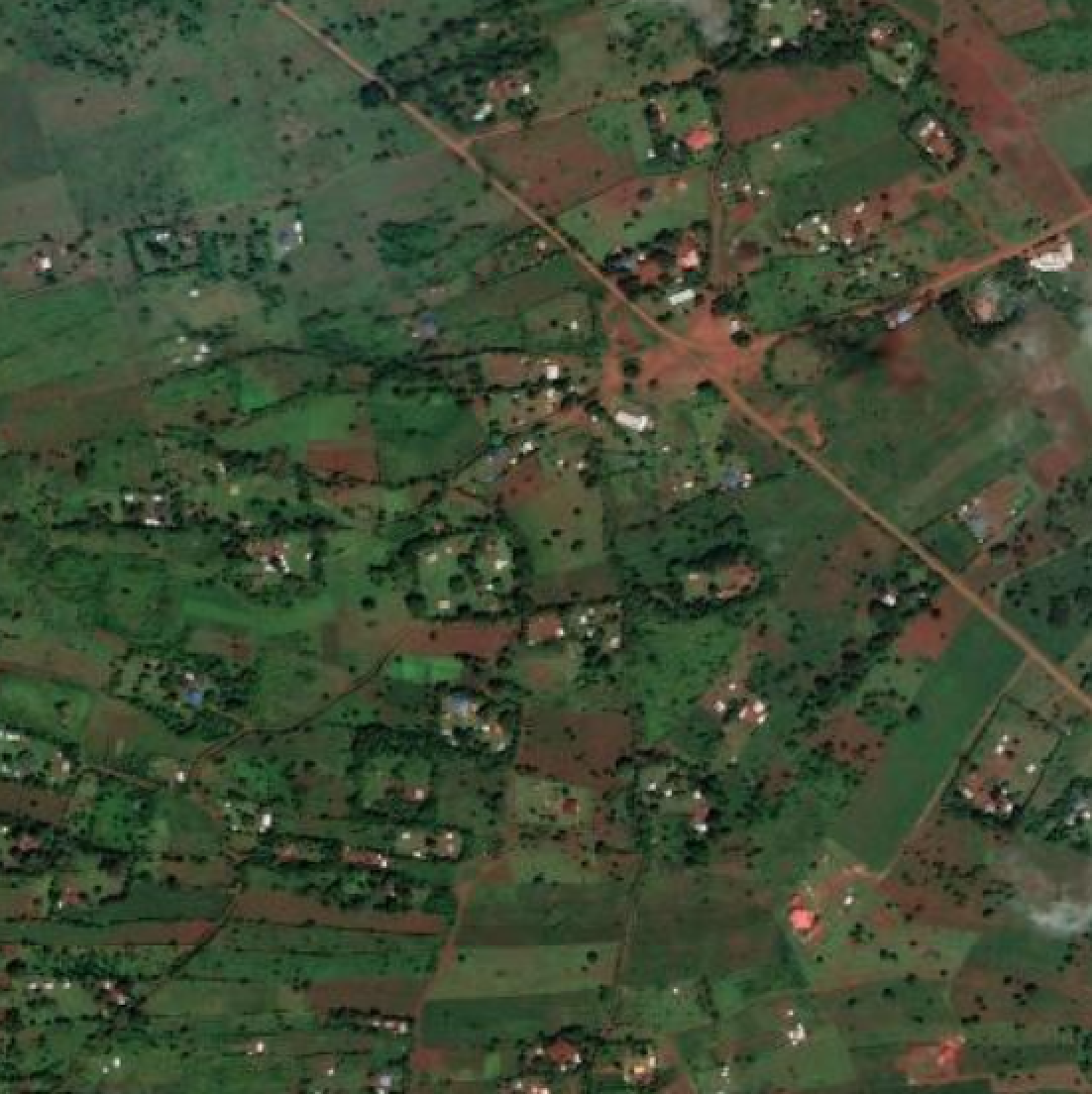}
        \caption{Kenya}
    \end{subfigure}
        \begin{subfigure}[b]{0.3\linewidth}
        \centering
        \includegraphics[width=\textwidth]{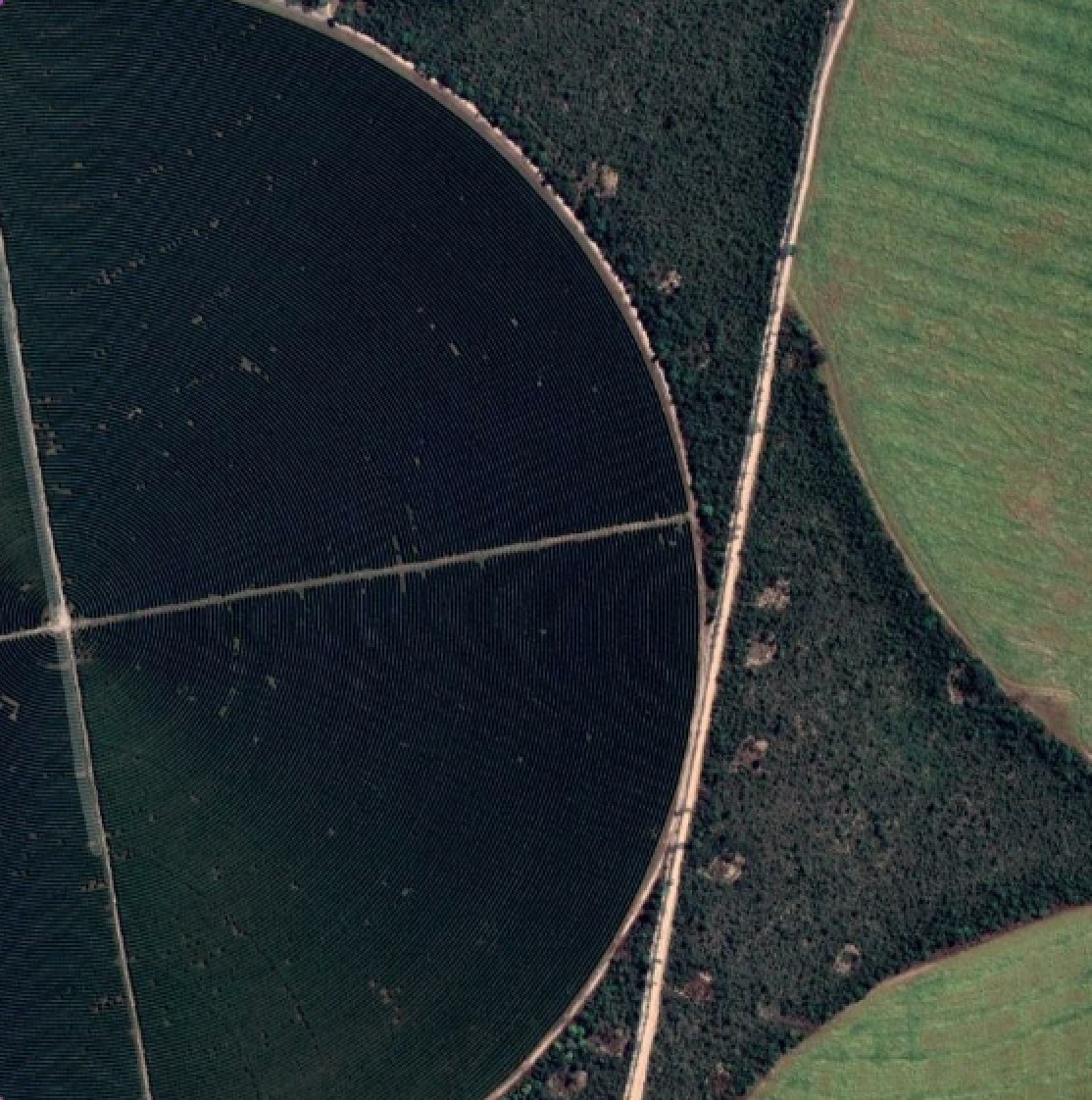}
        \caption{Brazil}
    \end{subfigure}
    \caption{Example 1km $\times$ 1km satellite images of the evaluation regions, demonstrating the variety in field sizes and agroecologies being evaluated. (Images were obtained from Google Earth Pro basemaps comprised primarily of high resolution Maxar images, and are reproduced with permission from \cite{cropharvest})}
    \label{fig:benchmarks}
\end{figure}

For the \textbf{crop type clasification} dataset, all LSTM-based classifiers were fine-tuned on the evaluation tasks for 250 gradient steps with batches containing 10 positive and 10 negative examples (as in \citet{cropharvest}). We show the variety of agro-ecologies represented in the crop type classification evaluation tasks in Figure \ref{fig:benchmarks}.

For the \textbf{yield estimation} dataset, all models were fine-tuned on each county for 15 gradient steps, with batches of size 10. The reduced fine-tuning steps relative to the crop classification dataset is due to the much lower amount of data available for each county (compared to the crop classification evaluation tasks). Some counties did not have any fine-tuning data available -- the results for these zero-shot counties are shared in Appendix \ref{ap:zero_shot}.

\subsection{Forgetfulness} \label{ap:forgetfulness}
We use the following thresholds to define task-memorization:
\begin{itemize}
    \item \textbf{Crop Type Classification}: An AUC ROC of 0.95 or above
    \item \textbf{Yield Estimation} An RMSE of 4 or less
\end{itemize}
In both cases, a training task was forgotten if it met the threshold for forgetfulness continuously over the last 20 epochs.

For the crop type classification, we note that the training batches were balanced to contain 10 positive and 10 negative examples, making AUC ROC appropriate.

\subsection{Task augmentation for geospatial MAML}
Defining tasks according to their geospatial boundaries allows for a form of weak task augmentation, by including nearby datapoints which are not explicitly within the boundary. For example, using a rectangular bounding box instead of a polygon when defining a political boundary includes nearby points which may not be inside the polygon. Similarly, for the yield estimation dataset we include nearby counties in tasks for MAML and TIML.

\section{Zero-shot learning} \label{ap:zero_shot}
For the \textbf{Yield estimation} task, some counties did not appear in the training data but were present in the evaluation data (i.e.~if the first year of data for a county is 2011, then there will be no training data for that county for the evaluation year 2011).

For these counties, the model is therefore evaluated in a zero-shot learning regime (the county is not present when training the meta-model, or during fine-tuning).

We record the results of the yield model in a zero-shot learning regime below in Table \ref{tab:zero_shot}. These results are included in the overall results reported in Table \ref{tab:yield_results}.

We highlight that very few counties are in this zero-shot regime, but include these results for completeness.

\begin{table}[]
    \centering
    \begin{tabular}{lrrrrr}
    \toprule[1.5pt]
    Model & 2011 & 2012 & 2013 & 2014 & 2015\\
    \midrule
    \# counties & 7 & 9 & 5 & 6 & 5 \\
    \midrule
         LSTM + TIML & 8.99 & 12.93 & 17.19 & 9.97 & 11.22  \\
         CNN + TIML & 10.44 & 7.02 & 9.81 & 7.25 & 11.89 \\
    \bottomrule
    \end{tabular}
    \caption{Zero-shot learning results: RMSE of the TIML model when measured only on counties not present during training (or fine-tuning). We note that these results were obtained with no training data about the county, in a zero-shot learning regime. The number of counties being tested is additionally recorded.}
    \label{tab:zero_shot}
\end{table}


\end{document}